\documentclass{article}

\usepackage{arxiv}

\RequirePackage{amsmath} 

\usepackage[utf8]{inputenc}
\usepackage[T1]{fontenc}

\usepackage{natbib}

\usepackage{hyperref} 
\usepackage{url}
\usepackage{amsfonts}
\usepackage{nicefrac}
\usepackage{microtype}     
\usepackage{cleveref}      
\usepackage{doi}
\usepackage{subcaption}
\usepackage{algorithm}
\usepackage[noend]{algpseudocode}
\usepackage{graphicx}
\usepackage{dsfont}

\usepackage{amsmath,amssymb,amsfonts}
\usepackage{dsfont}
\usepackage{subcaption}
\usepackage{graphicx}
\usepackage{url}
\usepackage{booktabs}

\usepackage{algorithm}%
\usepackage{algpseudocode}%

\usepackage{hyperref}
\usepackage{orcidlink}

\newtheorem{definition}{Definition}
\newtheorem{condition}{Condition}

\algdef{SE}[SUBALG]{Indent}{EndIndent}{}{\algorithmicend\ }%
\algtext*{Indent}
\algtext*{EndIndent}

\usepackage{esvect}


\makeatletter
\def\BState{\State\hskip-\ALG@thistlm}
\makeatother

\title{C-SHAP for time series: An approach to high-level temporal explanations}
\date{}

\usepackage{authblk}

\newbox{\orcid}\sbox{\orcid}{\includegraphics[scale=0.06]{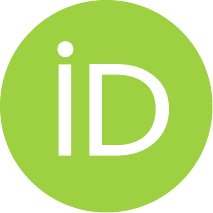}} 
\author[1,2]{%
	\href{https://orcid.org/0009-0004-9322-0674}{\usebox{\orcid}\hspace{1mm}Annemarie~Jutte\thanks{\texttt{a.m.p.jutte@saxion.nl}}}%
}
\author[1,2]{%
	\href{https://orcid.org/0000-0002-2760-6892}{\usebox{\orcid}\hspace{1mm}Faizan~Ahmed}%
}
\author[1]{%
	\href{https://orcid.org/0000-0002-2626-1837}{\usebox{\orcid}\hspace{1mm}Jeroen~Linssen}%
}
\author[2]{%
	\href{https://orcid.org/0000-0003-2436-1372}{\usebox{\orcid}\hspace{1mm}Maurice~van~Keulen}%
}
\affil[1]{Saxion University of Applied Sciences, Enschede, The Netherlands}
\affil[2]{University of Twente, Enschede, The Netherlands}

\renewcommand{\headeright}{}
\renewcommand{\undertitle}{}
\renewcommand{\shorttitle}{C-SHAP for time series}

\hypersetup{
pdftitle={C-SHAP for time series},
pdfauthor={A.~Jutte et al.},
pdfkeywords={Explainable AI, Time series, Concept-based XAI, SHAP},
}

\begin{document}
\maketitle

\begin{abstract}
In high-stakes domains, such as healthcare and industry, the explainability of AI-based decision-making has become crucial. Without insight into model reasoning, the reliability of these models cannot be ensured. Applications often rely on the time series data type which, unlike the image data type, is underexplored with respect to the development of explainable AI (XAI) techniques. Most existing XAI techniques for time series are focused on point- or subsequence-based explanations. This limits their usability since points and subsequences do not capture all relevant patterns and may not result in human-interpretable explainability. In this paper, we close this gap and propose a concept-based XAI approach (C-SHAP), where concepts are defined as high-level patterns extracted from the time series data. C-SHAP leverages the SHAP method to determine the influence of these concepts on predictions. The effectiveness of the developed framework is illustrated for use cases from healthcare and industry, in the form of Human Activity Recognition (HAR) and predictive maintenance.

\end{abstract}

\section{Introduction}
\label{sec:introduction}

As the adoption of artificial intelligence (AI) systems becomes increasingly prominent in various domains, awareness of the opaque nature of many of the underlying models is becoming more widespread~\citep{schwalbe_comprehensive_2023}. This lack of transparency in these `black box' models prevents the reliable use of AI and can lead to user distrust. To increase trust in AI models, users need insight into the models' reasoning~\citep{gunning2019xai}. The field of explainable AI (XAI) aims to uncover the reasoning of AI models by providing explanations~\citep{adadi_peeking_2018}.

This paper considers XAI for time series. AI has been successfully applied in a wide range of tasks involving time series, for example, in domains such as healthcare~\citep{morid_time_2023} and industry~\citep{jan_artificial_2023}. Traditionally, most XAI methods for time series are point- or subsequence-based~\citep{theissler_explainable_2022}. Point-based approaches determine the contribution of individual points in time towards an AI model's prediction~\citep{schlegel2019towards,wang2017time}. Subsequence-based approaches determine the contribution of groups of, generally sequential, points towards a model's prediction~\citep{cho2021interpreting,nayebi_windowshap_2023}. Both approaches offer valuable insight into a model's reasoning regarding local patterns. However, a model might not only leverage local patterns, but also higher-level patterns for decision making. Consider, for example, the global trend of a signal. The trend does not manifest itself in a single point or a subsequence of points, rather, it manifests itself globally across the points in the signal. Consequently, the attribution of the feature towards model reasoning is spread over points, and its importance cannot be expressed in terms of the attribution of individual points or subsequences. Hence, this paper will consider an alternative approach, providing attribution scores in terms of high-level features such as the trend of a signal, changes in variance, and its frequency components. 
 
These high-level features can be referred to as `concepts'. While the term `concept' is often used in literature on XAI~\citep{poeta_concept-based_2023}, an agreed-upon definition does not yet exist. In this paper, we follow the interpretation by Goyal et al.~\citep{goyal_explaining_2020}, of a concept as a ``higher level unit than low-level individual input features''. For time series, it has been shown that in some cases, concepts such as trend and frequency may better explain the behaviour of a model than point-based attribution~\citep{kusters_conceptual_2020}. In addition to the technical possibilities of concept-based explanations, human interpretable high-level concepts may better align with user understanding~\citep{poeta_concept-based_2023}.

However, technological advancement to detect and explain these high-level concepts is lagging behind. Our proposed method is a step in this direction, where we explain a model's predictions using high-level concepts that are understandable for domain experts or lay users. To this end, we present C-SHAP for time series. C-SHAP is based on SHapley Additive exPlanation (SHAP)~\citep{lundberg_unified_2017}, which is a post-hoc model-agnostic method that measures the contribution of features to model predictions. Whereas standard practice defines these features as low-level model input features, we define these features as concepts. 

An example of a C-SHAP explanation can be found in Figure~\ref{fig:local_turbo_cshap}. Using C-SHAP. We offer the explanation in terms of the attribution scores found for different concepts. The C-SHAP explanation uncovers the importance of the concept `Trend' to the model's prediction. For comparison, an explanation in terms of point-based attribution scores (using KernelSHAP~\citep{lundberg_unified_2017}) is shown in Figure~\ref{fig:local_turbo_kernel}. This point-based explanation provides little insight due to the spread of attribution across points.

We present two variants of C-SHAP: a fully post-hoc approach and a partially ante-hoc approach. The former approach assumes a concept-agnostic model trained without regard for the concepts under inspection, as illustrated in the top workflow in Figure~\ref{fig:cshap_approaches}. The latter assumes a concept-informed model trained directly on the concepts, as illustrated in the bottom of Figure~\ref{fig:cshap_approaches}.

\begin{figure*}[ht]
    \centering
    \includegraphics[width=\linewidth]{jutte1.png}
    \caption{In this paper, we present two approaches to concept-based explanations for time series using SHAP. One fully post-hoc concept-agnostic approach (top) and one partially post-hoc approach relying on a concept-informed model (bottom). The latter model is trained on the concepts under inspection. The approaches rely on the construction of high-level features (concepts) from time series. The attribution of concepts is determined using SHAP, where attribution scores are calculated by masking concepts.}
    \label{fig:cshap_approaches}
\end{figure*}

The fully post-hoc approach is specifically suited for use cases where modularity is desired, for example, in MLOps pipelines~\citep{ruf2022aspects}. This approach does not require any modifications to the model being explained, facilitating a seamless introduction in such pipelines. An advantage of the latter approach is that model performance could be increased by explicitly providing the model with concepts. Since the concepts are assumed to be important for model reasoning, steering the reasoning in their direction could be beneficial. For example, training deep learning models on time series decompositions has been shown to increase model performance in certain applications~\citep{liu2013forecasting,yang2019hybrid}.

For both approaches, concepts need to be constructed from the time-series data. In this paper, we construct concepts using time series decomposition, due to its straightforward interpretation and implementation. We demonstrate the effectiveness of C-SHAP using the Discrete Wavelet Transform (DWT) decomposition~\citep{heil1989continuous}. Additionally, we present and demonstrate the effectiveness of a custom decomposition approach, designed for human-interpretability.

The C-SHAP approach for concept-agnostic models is similar to approaches generating SHAP values in the frequency domain using Fourier filtering~\citep{herwig2023explaining,brusch2024flextime}. However, we expand this approach by including features from both the time and frequency domains. The C-SHAP approach for concept-informed models is a formalisation and generalisation of methods in literature which apply SHAP to models trained on decomposition components~\citep{han2024comparison,cu2025time}.

In this paper, we conceptualise the C-SHAP generalisation of both approaches. Furthermore, we mitigate the gap between theory and practice by applying C-SHAP to use cases from two different domains: Human Activity Recognition (HAR) and predictive maintenance. Specifically, we apply C-SHAP to a locomotion classification task (HAR), and a remaining useful life prediction task (predictive maintenance). We aim to illustrate how C-SHAP can provide user-friendly explanations for end-users and domain experts.

Summarising, this research makes the following contributions to the state of the art:
\begin{enumerate}
    \item The formalisation of two approaches to concept-based explanations for time series using SHAP: one fully model-agnostic post-hoc approach and one approach relying on a concept-informed model.
    \item A demonstration of the use of time series decomposition in concept-based explanations.
    \item The design of a custom time series decomposition algorithm aimed at human-centred interpretability.
    \item The application of C-SHAP to use cases from the Human Activity Recognition and predictive maintenance domains.
\end{enumerate}

The remainder of this article is structured as follows: in Section~\ref{sec:related_work} we provide an overview of the state of the art, in Section~\ref{sec:conceptualisation} we provide the formal conceptualisation of C-SHAP, in Section~\ref{sec:implementation} we describe how we have implemented C-SHAP using time series decomposition, in Section~\ref{sec:experimental} we discuss the experimental setup for the application of C-SHAP, and in Section~\ref{sec:results} we present the resulting explanations.

\section{Related work}
\label{sec:related_work}
A substantial portion of literature on XAI for time series is centred around so-called point-based explanations~\citep{theissler_explainable_2022}. Point-based explanations provide relevance scores to individual points in a series. Relevance scores can be inherently implemented into models for time series, for example using attention mechanisms~\citep{vaswani2017attention} in recurrent neural networks (RNNs)~\citep{karim2017lstm,li2019ea}. Another option, not requiring any modifications to the models, is the use of post-hoc methods to calculate the attribution of points. These attribution-based methods are either model-specific, such as gradient-based methods~\citep{sundararajan2017axiomatic,wang2017time} or model-agnostic, such as the application of the popular SHAP~\citep{lundberg_unified_2017} or LIME~\citep{ribeiro2016should} algorithms to time series~\citep{schlegel2019towards,meng2024segal}.

SHAP~\citep{lundberg_unified_2017} is a post-hoc XAI method which determines the contribution of features towards model predictions. To this end, SHAP measures the change in model prediction when features are excluded from the data. To take cooperation between features into account, SHAP considers the effect when including or excluding features in coalitions: groups of features `working together'. We will discuss the details of SHAP in Section~\ref{sec:conceptualisation}. To generate a point-based explanation, SHAP can be directly applied to a time series, when defining each point in the series as a feature.

A disadvantage of SHAP is its computational cost. SHAP considers all possible coalitions of features. As a consequence, the computational cost of SHAP increases exponentially with the number of features. Since point-based XAI approaches for time series consider each point in time to be an individual feature, this computational complexity is specifically limiting for long time series. KernelSHAP~\citep{lundberg_unified_2017} mitigates this approach by randomly sampling a subset of coalitions to consider. However, to obtain qualitative results, KernelSHAP may remain computationally expensive. Additionally, KernelSHAP does not take the sequentiality of recurrent models into account~\citep{bento_timeshap_2021}. 

TimeSHAP~\citep{bento_timeshap_2021}, is a SHAP-based approach specifically designed for time series. TimeSHAP operates under the assumption that RNNs mainly rely on the latter points of a sequence, and reduces computational cost by combining the initial points into a single feature. This assumption may not be true for all use cases. WindowSHAP~\citep{nayebi_windowshap_2023} instead mitigates issues with KernelSHAP for time series by considering subsequences of points, windows, rather than individual points. 

While point- or subsequence-based approaches can uncover relevant details regarding model reasoning, they are limited to explanations in terms of local patterns. High-level approaches, using so-called `concepts', can expand the XAI toolbox. Recently, concept-based explanations have become popular across data types~\citep{poeta_concept-based_2023}. Testing with Concept Activation Vectors (TCAV) \citep{kim_interpretability_2018} is a leading approach in concept-based XAI, also for time series~\citep{mincu_concept-based_2021,brenner_concept-based_2024}. TCAV provides explanations based on Concept Activation Vectors, representing a model's response to predefined concepts.  ConceptSHAP~\citep{yeh_completeness-aware_2020} combines TCAV and SHAP. A completeness score is used to evaluate the concepts used in TCAV. 

A disadvantage of TCAV is that it relies on manually annotated examples and counterexamples of concepts. The collection of such an annotated dataset can be a labour-intensive task. Similarly, Concept Bottleneck Models (CBMs)~\citep{koh2020concept} are interpretable-by-design models which explicitly link a model's activations in the bottleneck layer to concepts, generally using annotated datasets.

Sun et al.~\citep{sun_explain_2024} take a different approach, where they define a concept as an image segment. They apply SHAP to automatically extracted image segments. This is similar to the approach presented in this article. We define concepts using mathematical constructs, as will be demonstrated in Section~\ref{sec:implementation}. Through this approach, C-SHAP does not require annotated concepts, yet still provides control over the concepts used in the explanation.

Our approach for concept-agnostic models is similar to methods that use localized attribution to provide explanations in the frequency domain~\citep{herwig2023explaining,brusch2024flextime}. Local features in the frequency domain correspond to global features in the time domain, i.e. concepts. However, unlike these methods, C-SHAP is not restricted to either the time or frequency domain, C-SHAP can combine the two. For example, the custom decomposition presented in Section~\ref{sec:sensor_decomposition}, includes a component capturing the low-frequency behaviour of the signal, and also a component capturing changes in amplitude over time.

Our decomposition-based approach also relates to existing work that embeds time series decomposition in model training and applies feature-attribution to determine component attribution~\citep{han2024comparison,cu2025time,cleveland1990stl}. We generalize these SHAP-based methods by formalizing C-SHAP and explicitly relating it to concept-based XAI. Additionally, while state-of-the-art approaches are typically application-specific, we demonstrate the wider applicability of the C-SHAP approach.

\section{Conceptualisation}
\label{sec:conceptualisation}
Shapley values~\citep{shapley_value_1952} are an approach from game theory to determine the contribution of variables to a function output. Given a black box data-driven model, SHAP~\citep{lundberg_unified_2017} is a method that estimates Shapley values for data features as their contribution to the model output. These estimated Shapley values are called SHAP values. In this section, we discuss SHAP and its extension C-SHAP for time series, which determines the SHAP values of concepts. For an overview of the C-SHAP approach, see Figure~\ref{fig:cshap_approaches}. We start this section by expanding the definition of SHAP values. We then discuss the required concept construction and concept masking algorithms for both concept-informed and concept-agnostic models. We finish by imposing a condition on the concept construction to ensure valid SHAP values.

\subsection{SHAP values for C-SHAP}
For univariate time series, consider an input sample $\textbf{y}(t) = \left(x_1, x_2, ..., x_n\right)\allowbreak\in\mathds{R}^{n}$. SHAP values are based on coalitions, if $G = \{g_1, g_2, ..., g_m\}$ is the full set of features, a coalition is defined as a subset of features ($S\subseteq G$). For each coalition $S$, a coalition vector $\mathbf{z}_S \in \{0, 1\}^m$ is defined, where $z_i = 0$ if $g_i \notin S$ and $z_i = 1$ if $g_i \in S$. 

To determine the SHAP values, the model $f$ under inspection is applied to a masked model input $h_\mathbf{y}(\mathbf{z}_S)$, where any features in $\mathbf{y}(t)$ which are not in coalition $S$ are masked, i.e. `toggled off'. Features can be masked by replacing their value with an uninformative background value. We denote the model output of the masked input as $f_\mathbf{y}(\mathbf{z}_S) = f(h_\mathbf{y}(\mathbf{z}_S))$. 

The SHAP value $\phi_i$ for feature $g_i$ is calculated as a weighted average, over all coalitions, of the difference in model output when $g_i$ is either included or excluded in the coalition~\citep{lundberg_unified_2017}:

\begin{equation}
    \phi_i(f, \mathbf{y}) = \sum_{S \subseteq G\backslash{g_i}} \frac{|S|(m - |S| - 1)!}{m!} \left[f_\mathbf{y}(\mathbf{z}_{S\cup g_i}) - f_\mathbf{y}(\mathbf{z}_{S})\right].
    \label{eq:shap}
\end{equation}

Previous methods applying SHAP to time series consider features $g_i$ to be data points or subsequences~\citep{bento_timeshap_2021,nayebi_windowshap_2023}. For C-SHAP we follow \cite{sun_explain_2024} and consider a set of high-level concepts $C$ as features, i.e. $G = C$. 

We define a concept $c_i$ as a feature of a higher level of abstraction than input features~\citep{goyal_explaining_2020}. Furthermore, we define $C = \left\{c_1, c_2,\cdots, c_m\right\}$ to be a set of concepts. Consequently, coalition $S$ is a subset of concepts, i.e. $S \subseteq C$. Then, we can adapt (\ref{eq:shap}) by substituting $G = C$, and it follows that the SHAP values $\phi_i$ for concept $c_i$ are given by:
\begin{equation}
    \phi_i(f, \mathbf{y}) = \sum_{S \subseteq C\backslash{c_i}} \frac{|S|(m - |S| - 1)!}{m!} \left[f_\mathbf{y}(\mathbf{z}_{S\cup c_i}) - f_\mathbf{y}(\mathbf{z}_{S})\right].
    \label{eq:conceptbshap}
\end{equation}
Note that since we follow (\ref{eq:shap}), the properties of local accuracy, missingness and consistency~\citep{lundberg_unified_2017} are satisfied.

To obtain the SHAP values for C-SHAP using (\ref{eq:conceptbshap}), we need a concept construction algorithm to obtain the set of concepts $C$ and we need a concept masking algorithm to obtain the masked model output $f_\mathbf{y}(\mathbf{z}_{S})$. For SHAP, features are masked by setting them to an uninformative value. Point-based attribution methods mask features by changing the values of specific points or segments~\citep{bento_timeshap_2021,nayebi_windowshap_2023}. C-SHAP instead masks concepts $c_i$.

For C-SHAP we consider two types of models that can be trained, concept-agnostic and concept-informed models, see Figure~\ref{fig:cshap_approaches}. A concept-agnostic model is trained using the raw model input $\mathbf{y}(t)$. A concept-informed model is trained directly on the concepts $C$, where each concept is passed to an input channel of the model. As will be discussed in the next section, the model type affects the concept masking. 

\subsection{Concept construction and masking}

The concept construction algorithm does not depend on the choice of a concept-agnostic or a concept-informed model. For the concept construction, to obtain the set of concepts $C$, a function $d(\cdot)$ needs to be selected which extracts the set of concepts $C=\{c_1, c_2,\cdots, c_m\}$ from an input sample $\mathbf{y}(t)$, i.e. $C=d(\mathbf{y}(t))$. In the next section, we will pose a condition on this function. Note, for the concept-informed model, the concepts do need to be constructed before training. For the concept-agnostic model, all C-SHAP steps can be executed in a post-hoc manner.

The concept masking does depend on the choice for a concept-agnostic or concept-informed model. For the concept-informed model, masked concepts can be passed directly to the model. For the concept-agnostic model, the masked concepts should be aggregated into a univariate signal in which the concepts are toggled off, see Figure~\ref{fig:cshap_approaches}.

Concretely, for the concept-informed approach, the masked model input is given by:
\begin{equation}
    h^{\text{informed}}_\mathbf{y}(\mathbf{z}_S) = \left[h_1, h_2,\cdots, h_m\right],
    \label{eq:masking_def}
\end{equation}
where, given an uninformative replacement concept $\bar{c}_i$ for $c_i$:
\begin{equation*}
    h_i=\begin{cases}
        c_i & \text{if }  i \in S,  \\
        \bar{c}_i & \text{if } i \notin S.
    \end{cases}
\end{equation*}
A demonstration of how this masking is implemented will be given in Section~\ref{sec:component_masking}.

For the concept-informed approach, the model input should be a univariate signal. Hence, we require a function $r: \mathds{R}^{m\times n} \to \mathds{R}^{n}$ that recombines the masked concepts:
\begin{equation}
        h^{\text{agnostic}}_\mathbf{y}(\mathbf{z}_S) = r(h_1, h_2, \cdots, h_m).
        \label{eq:masked_agnostic}
\end{equation}

\subsection{Conditions}
In this section we pose a condition on the concept construction algorithm. The masked model input $h_\mathbf{y}(\mathbf{z}_S)$ for coalition $S$ represents the sample $\mathbf{y}(t)$ where any concepts $c_i\notin S$ are removed. This means that if we have a coalition $C = S$, including all concepts in $\mathbf{y}(t)$, we need $h_\mathbf{y}(\mathbf{z}_S) = \mathbf{y}(t)$. We formalize this in the following condition.

\begin{condition}
Given coalition $S = C$, representing the full set of concepts, the masked model input $h_\mathbf{y}(\mathbf{z}_C)$ needs to be equal to the model input. For concept-agnostic models this means $h^{\text{agnostic}}_\mathbf{y}(\mathbf{z}_C) = \mathbf{y}(t)$. For concept-informed models this means $h^{\text{informed}}_\mathbf{y}(\mathbf{z}_C) = [c_1, c_2,\cdots, c_m]$.
\label{cond:equal}
\end{condition}

For the concept-informed approach, if we have coalition $S = C$, we have $h_i = c_i$ for $i\in\{1, 2, \cdots, m\}$. Hence, it follows from (\ref{eq:masking_def}) that $h^{\text{informed}}_\mathbf{y}(\mathbf{z}_S) = \left[c_1, c_2, \cdots, c_m\right]$. Therefore, Condition~\ref{cond:equal} will be satisfied regardless of the chosen construction and masking algorithms.

However, for the concept-agnostic approach, Condition~\ref{cond:equal} poses a specific condition on $r$ in (\ref{eq:masked_agnostic}). For coalition $S=C$, we again have $h_i =c_i$ for $i\in\{1, 2, \cdots, m\}$. Furthermore, for the masked model input we have, from (\ref{eq:masked_agnostic}):
\begin{align*}
h_\mathbf{y}^{\text{agnostic}}(\mathbf{z}_C) = r(h_1, h_2, \cdots, h_m) =r(c_1, c_2, \cdots, c_m)= r(d(\mathbf{y}(t))),
\end{align*}
where $d(\cdot)$ is the concept construction function. Hence, by Condition~\ref{cond:equal}, we require:
\begin{equation*}
    r(d(\mathbf{y}(t))) = \mathbf{y}(t).
\end{equation*}
Therefore, $r$ needs to be the inverse of $d$. As a consequence, we pose the following condition on concept construction.

\begin{condition}
For C-SHAP for concept-agnostic models, the concept construction function $d(\cdot)$ needs to be invertible.
\label{cond:invertible}
\end{condition}

In the next section, we present an implementation for concept construction, satisfying Condition~\ref{cond:invertible}, and we present an implementation for concept masking. Our concept construction implementation is based on time series decomposition. Time series decomposition approaches are generally invertible, since invertibility ensures recoverability of the original time series. Note that C-SHAP is not limited to time series decomposition. Future research should explore other concept construction approaches.

\section{Implementation}
\label{sec:implementation}
In this section, we present an implementation of the conceptualisation of C-SHAP. As previously discussed, we need to select an algorithm for concept construction and an algorithm for concept masking. Additionally, depending on the choice of a concept-agnostic or concept-informed model (see Figure~\ref{fig:cshap_approaches}), model training is explicitly part of the explanation process. 
For the concept construction, we use identical implementations for the concept-agnostic and -informed approaches, while the concept masking algorithm is slightly different for the two.

\subsection{Concept construction}
\label{sec:concept_construction}
For this implementation, we consider concept construction through time series decomposition. By Condition~\ref{cond:invertible}, for the C-SHAP approach relying on a concept-agnostic model, the concept construction needs to be invertible. Time series decomposition offers this invertibility by its definition in terms of component aggregation.

\begin{definition}
Given a time series $\mathbf{y}(t) \in \mathds{F}^n$, where $\mathds{F}$ is a (real or complex) field, we define its decomposition into a set of components $ C = \{\mathbf{y}_1(t), \mathbf{y}_2(t),\allowbreak \cdots, \allowbreak\mathbf{y}_n(t):\mathbf{y}_i(t) \in \mathds{F}^n\}$ such that:
\begin{equation}
    \mathbf{y}(t) = \mathop{\otimes}_{i = 1}^n \alpha_i \hspace{1mm}\mathbf{y}_i(t),
    \label{eq:general_decomposition}
\end{equation}
where $\otimes$ is an aggregation operator and $\alpha_i$ are the component weights. For this implementation of C-SHAP, we define the set of components as our concept set $C$ in~(\ref{eq:conceptbshap}). 
\end{definition}

In this research, different decompositions are explored. The most popular approach for time series decomposition is arguably the Fourier transform, which separates a signal into global frequency components~\citep{bracewell1989fourier}. While concept construction relying on the Fourier transformation is applicable to C-SHAP and can be beneficial for certain applications, we argue it would be a limiting decision for the demonstration in this paper. Fourier transforms are not able to localise frequencies in time. As a consequence, Fourier decompositions might result in physically non-interpretable components if the data is not strictly periodic or stationary~\citep{huang1998empirical}. We do not want to put these restrictions on the data, therefore we turn to different decomposition approaches. Namely, we consider the Discrete Wavelet Transform (DWT) and a custom decomposition.

\subsubsection{Discrete Wavelet Transform}
The Discrete Wavelet Transform~\citep{heil1989continuous} relies on localised wavelet functions of different scales. Wavelets are a set of functions that adhere to specific properties~\citep{heil1989continuous}. By sliding wavelets of different scales along a time series, DWT can extract local frequency components. This localisation makes DWT appropriate for non-periodic signals. However, by Heisenberg's uncertainty principle~\citep{heisenberg1927anschaulichen}, DWT loses resolution in the frequency domain due to this detailed localization~\citep{chui1992introduction}. Therefore, rather than extracting components of specific frequencies, DWT extracts components of different `levels of detail'.

More concretely, the Discrete Wavelet Transform (DWT) splits a signal into an `Approximation' and `Detail' component using high and low-pass filters. The DWT can be iteratively applied to the `Detail' signal to obtain more components. We denote the DWT decomposition of detail level $N_\text{Detail}$ as follows:
\begin{equation}
    \mathbf{y}(t) = \mathbf{y}_\text{Approximation}(t) + \sum_\text{i = 1}^{N_\text{Detail}} \mathbf{y}_\text{Detail-$i$}(t).
    \label{eq:DWT_decomposition}
\end{equation}
Using this decomposition, we can define concepts for our explanations. We define the concept `Approximation', and the concepts `Detail-i', for $i \in \{1, \cdots, N_\text{Detail}\}$.

\subsubsection{Custom decomposition}
~\label{sec:sensor_decomposition}
While DWT is used in technical applications, its interpretation may still be challenging, specifically for non-experts. Therefore, we also present an alternative custom decomposition. Rather than defining mathematical properties and relating these to real-world phenomena, we start by defining concepts meaningful to humans and relating these to mathematical properties that can be used to extract them as components. The defined concepts can be found in Table~\ref{tab:custom_decomposition}.

\begin{table*}[!htp]
\centering
\caption{Concepts defined for the custom decomposition. Per concept, its definition and the mathematical notation of the component are listed. Additionally, it is indicated whether the corresponding component is additive ($+$) or multiplicative ($\times$).}
\label{tab:custom_decomposition}
\begin{tabular}{@{}lclc@{}}
\toprule
Concept        & Notation & Definition                                           & Operation      \\ \midrule
Trend          &     $\mathbf{y}_\text{Trend}(t)$    & The increase or decrease of a time series over time. & $+$       \\
Bias           &     $\mathbf{y}_\text{Bias}(t)$     & The average level of a time series.                  & $+$       \\
Scale          &      $\mathbf{y}_\text{Scale}(t)$    & The scale of the amplitude of a time series.         & $\times$ \\
Variance       &     $\mathbf{y}_\text{Var}(t)$     & Changes in amplitude over time in a time series.     & $\times$ \\
Low frequency  &     $\mathbf{y}_\text{LF}(t)$     & Slow changes over time in a time series.             & $+$       \\
High frequency &     $\mathbf{y}_\text{HF}(t)$     & Rapid changes over time in a time series.            & $+$       \\ \bottomrule
\end{tabular}
\end{table*}

In contrast to the DWT, which is additive, the custom decomposition is hybrid, meaning it has both additive as well as multiplicative components, see Table~\ref{tab:custom_decomposition}. This makes the decomposition order-dependent. We define the following ordering in the decomposition:
\begin{equation}
     \mathbf{y}(t) = \mathbf{y}_\text{Trend}(t) + \mathbf{y}_\text{Bias}(t) \nonumber
     + \mathbf{y}_\text{Scale}(t) \times (\mathbf{y}_\text{LF}(t) + \mathbf{y}_\text{Var}(t) \times \mathbf{y}_\text{HF}(t)).
\label{eq:custom_decomposition}
\end{equation}
While this ordering might not be the only valid option, it was chosen for pragmatic reasons. The order defines how components should be extracted. First extracting the components for the `Trend' and `Bias' concepts simplifies the extraction of the wave-based concepts. For these concepts, `Scale' describes the amplitude of the remaining signal, `Variance' describes the changes in the amplitude of the `High frequency' part of the signal. Further research could explore the effects of alternative orderings.

Using the definitions of the concepts, we propose algorithms to extract them as components. We emphasise that the proposed decomposition is a heuristic-based approach. The choices made for this decomposition are pragmatic and may not be unique. For example, for the definitions of the concepts `Low frequency' and `High frequency', the difference between slow and rapid changes is subjective. Do note that the proposed decomposition is deterministic. When fixing the hyperparameters, the decomposition will consistently yield the same components. We describe our decisions regarding the decomposition in the remainder of this section.

The components are extracted iteratively from $\mathbf{y}(t)$, matching their order in the equation, from left to right. The first component $\mathbf{y}_\text{Trend}(t)$ is extracted from $\mathbf{y}(t)$, then the next component is extracted from the residual signal $\mathbf{y}^{r_1}(t) = \mathbf{y}(t) - \mathbf{y}_\text{Trend}(t)$, and so on. After extracting the other components, including the `Low frequency' component, we define the `High frequency' component as the residual.

We will now discuss the extraction of the components. We assume `Trend' to be a linear line. Therefore, we define the component $\mathbf{y}_\text{Trend}(t)$ as the slope of the linear least squares fit to $\mathbf{y}(t)$. `Bias' is implemented as the average level of the signal, $\mathbf{y}_\text{Bias}(t) = \text{mean}(\mathbf{y}^{r_1}(t))$. We implement `Scale' as the maximum amplitude of the signal, $\mathbf{y}_\text{Scale}(t) = \max_t|\mathbf{y}^{r_2}(t)|$. Note, we implement `Scale' to be a multiplicative component. Hence, $\mathbf{y}^{r_3}(t) = \mathbf{y}^{r_2}(t) / \mathbf{y}_\text{Scale}(t)$.

The component for the `Low frequency' concept is implemented using DWT filtering to extract low frequency components from the signal $\mathbf{y}^{r_3}(t)$. As previously discussed, DWT decomposes a series into a low-pass approximation series $A$ and high-pass detail series $\{D_1, D_2,\cdots, D_k\}$. High absolute values in the detail series indicate a strong presence in the original series, low values are assumed to relate to noise. The set of detail series can be thresholded, and the result can be split into the `Low frequency' component $\mathbf{y}_\text{LF}(t)$ and detail $\mathbf{y}^{r_4}(t)$~\citep{donoho1994ideal}. The threshold hyperparameter influences how much detail is filtered from the signal.

To extract the component for the concept `Variance' from $\mathbf{y}^{r_4}(t)$, a sliding window approach is applied. Given a window size $w_\text{size}$, we slide a window over the signal. For each point $x_i$ in $\mathbf{y}^{r_4}(t)$, we calculate the standard deviation of the window of size $w_\text{size}$ with $x_i$ as the middle point. For the points where the window extends over the boundary of the series, we crop the window to the start or end of the series. To stabilise the result, a moving average is applied to the resulting standard deviation values, using the same window size $w_\text{size}$ and middle points $x_i$. The result is normalised through division by the maximum value, resulting in $\mathbf{y}_\text{Var}(t)$.

Finally, the component corresponding to the concept `High frequency' is defined as $\mathbf{y}_\text{High freq}(t) = \mathbf{y}^{r_5}(t)$, the residual after the other components are extracted. 

\subsection{Model training}
As discussed in Section~\ref{sec:conceptualisation}, the models used in C-SHAP can be trained without regard for the concepts under inspection (the concept-agnostic approach) or using the constructed concepts as input features (the concept-informed approach). For the concept-agnostic approach, a black box model is trained on the raw input data. For the concept-informed approach, a multi-channel model is trained, where each channel receives the component corresponding to a concept.

\subsection{Concept masking}
\label{sec:component_masking}
For the calculation of SHAP values, a masking algorithm is needed to obtain masked model output $f_\mathbf{y}(\mathbf{z}_S)$. In this section, we discuss an implementation of the approach described in Section~\ref{sec:conceptualisation}.

For SHAP, features are masked by replacing them with `uninformative' data. In our case, this means concepts should be replaced by `uninformative' concepts. There are multiple approaches to simulate uninformative data~\citep{shap_documentation}. In this paper, we use the approach where uninformative data is approximated by repeatedly replacing concepts with concepts sampled from the training data.

For the concept-informed approach, each input channel receives a concept, i.e. a component. Concepts are masked by replacing the associated components with their counterparts from the training data. Concretely, given input sample $\mathbf{y}(t)$ and a randomly selected sample from the training data $\mathbf{y}_b(t)$, we decompose the samples as $\mathbf{y}(t) = (\mathbf{y}_1(t), \mathbf{y}_2(t),\cdots, \mathbf{y}_m(t))$ and $\mathbf{y}_b(t) = (\mathbf{y}_1^b(t), \mathbf{y}_2^b(t),\cdots, \mathbf{y}_m^b(t))$. Based on~(\ref{eq:masking_def}), and given coalition vector $\mathbf{z}_S$, we define the masked model input as:
\begin{equation*}
    h^{\text{informed}}_\mathbf{y}(\mathbf{z}_S, \mathbf{y}_b(t)) = \left[\mathbf{h}_1, \mathbf{h}_2,\cdots, \mathbf{h}_m\right],
\end{equation*}
where:
\begin{equation*}
    \mathbf{h}_i=\begin{cases}
        \mathbf{y}_i(t) & \text{if }  i \in S,  \\
        \mathbf{y}_i^b(t) & \text{if } i \notin S.
    \end{cases}
\end{equation*}

To obtain an approximation of the masked model output, $N$ training samples $\{\mathbf{y}_{b_1}(t), \mathbf{y}_{b_2}(t),\cdots, \mathbf{y}_{b_N}(t)\}$ are sampled. This results in the following approximation of the masked model output:
\begin{equation*}
    f_\mathbf{y}(\mathbf{z}_S) \approx \frac{1}{N} \sum_{j=1}^Nf(h^{\text{informed}}_\mathbf{y}(\mathbf{z}_S, \mathbf{y}_{b_j}(t))).
\end{equation*}

This approach can be expanded to the concept-agnostic version, requiring one additional step where the masked components are composed into the required univariate model input form (cf.~(\ref{eq:general_decomposition})), see (\ref{eq:masked_agnostic}):
\begin{equation*}
    h^{\text{agnostic}}_\mathbf{y}(\mathbf{z}_S, \mathbf{y}_b(t)) = \mathop{\otimes}_{i = 1}^m \alpha_i\cdot \mathbf{h}_i.
\end{equation*}

\section{Experimental setup}
\label{sec:experimental}
To demonstrate and validate the C-SHAP implementation, we apply C-SHAP to black box models trained on two tasks: Human Activity Recognition (HAR) and remaining useful life estimation in predictive maintenance. These tasks were selected to showcase the generalizability of C-SHAP across domains. In this section, we describe the setup in our experiments.

\subsection{Data}
In this section, we detail the datasets used and their preparation for the experiments in this paper. Each of the datasets is normalised using the mean and the standard deviation of the training data.

\subsubsection{OPPORTUNITY}
The OPPORTUNITY dataset~\citep{roggen2010collecting} is a dataset for Human Activity Recognition(HAR). The dataset contains sensor data collected from four subjects undertaking sequences of daily activities. The dataset contains different types of annotations, we only consider the locomotion annotations. There are four classes of locomotion in the dataset: `Stand', `Walk', `Sit' and `Lie'.

For each subject, five runs of activities are available, four runs of daily activities (ADL) and one drill run. For subject~4, two ADL runs were distorted for additional challenge, these are not considered for our experiments. We use the ADL runs of subjects 2 and 3 as test data. All other non-excluded runs are used as training data. 

Multiple sensors were placed on the subjects. In our univariate experiments, we selected the y-axis of the accelerometer placed above the right knee since in pre-analysis it turned out to be most influential for the selected activities. Sensor measurements are available at a frequency of approximately 30Hz. 

Rows with missing values are removed. The remaining rows are divided into sets of consecutive rows. We segment the consecutive rows into samples of 90 data points (approximately 3s each). To increase the number of training samples, we use a sliding window approach, with an overlap of $50\%$ with the previous and next window. This results in 11,517 training and 1,967 testing samples.

\subsubsection{Turbofan}
The Turbofan dataset~\citep{saxena2008damage} contains data regarding a simulation model of an aircraft turbofan. Experiments were run where damage was introduced into the system and propagated over time. Experiments are run until the damage grows into failure. The task is to predict the remaining useful life of the machine in cycles.

Data is available for different types of degradation and environmental conditions. In this paper, only HPC degradation under sea level conditions is considered. A wide range of sensor measurements is available. For the univariate experiments in this paper, we use the P30 measurements: the total pressure at the HPC outlet. 

The data consists of training runs, which contain data until breakdown of the turbofan, and shorter test runs, that stop at some cycle before breakdown. We split the training runs into windows of 100 cycles, with an overlap of five cycles between windows. For each window, we set the number of cycles until breakdown as the target variable. For the test data, we trim all runs to 100 cycles, cutting off the start of longer runs. Runs that are shorter than 100 cycles are discarded. This results in 2,086 training samples and 70 testing samples.

\subsection{Decompositions}

\subsubsection{Discrete Wavelet Transform}
For the Discrete Wavelet Transform decomposition, we use the Daubechies wavelet with six vanishing moments. The decomposition is performed at a detail level of three, i.e. $N_\text{detail} = 3$ in (\ref{eq:DWT_decomposition}), resulting in four components. The PyWavelets package~\citep{lee2019pywavelets} is used for the implementation.

\subsubsection{Custom decomposition}
For the custom decomposition, we use the decomposition and ordering of components from (\ref{eq:custom_decomposition}). The decomposition requires a few hyperparameters. For the extraction of the $\mathbf{y}_\text{LF}(t)$ component, we use DWT-based filtering, as described in Section~\ref{sec:sensor_decomposition}. As previously for the DWT decomposition, the Daubechies wavelet with six vanishing moments is used. For the filtering of the detail coefficients, we use soft thresholding~\citep{donoho1994ideal} with a threshold of $T = \sigma \sqrt{2\cdot\log(n)}$, where $n$ is the length of our sample $\mathbf{y}(t)$ and we set $\sigma = 1$. For the sliding windows used to extract $\mathbf{y}_\text{Var}(t)$, we use a window size of 27 points for the OPPORTUNITY dataset and 30 points for the Turbofan dataset.

\subsection{Black box models}
For each dataset, black box models were developed using PyTorch~\citep{paszke2019pytorch}. We emphasise that C-SHAP is applicable regardless of the choice of model. In our experiments, we consider GRU-based models~\citep{cho_learning_2014}.

For the concept-informed approach, models were trained for the different decompositions. For the concept-agnostic approach, models were trained on the raw univariate samples. For all types, ten models were trained. The model architectures are the same for the concept-agnostic and -informed models, only the number of input channels differs.

Each model consists of multiple GRU layers followed by a single fully connected layer. Depending on the task, the model uses uni- or bidirectional GRU layers. For each model, all GRU layers have the same hidden size. For some models, dropout is used. The configuration of the models for each dataset can be found in Table~\ref{tab:model_architecture}.

The mean performance on the training and testing data for the different models can be found in Table~\ref{tab:model_performance}. For the OPPORTUNITY classification model, the performance is given as the accuracy, weighted over the classes, in percentage. For the Turbofan regression model, the performance is given as the RMSE in cycles.

\begin{table}[htp]
\centering
\caption{Configuration of the models trained on each dataset.}
\label{tab:model_architecture}
\begin{tabular}{lcccc}
\toprule
\textbf{Dataset} &
\textbf{Layers} &
\textbf{Hidden neurons} &
\textbf{Dropout} &
\textbf{Bidirectional} \\
\midrule
\textbf{OPPORTUNITY}  & 4 & 64 & 0.2 & True  \\
\textbf{Turbofan}     & 1 & 64 & 0.0 & False \\
\bottomrule
\end{tabular}
\end{table}

\begin{table*}[t]
\caption{Performance (mean $\pm$ std over 10 runs) for the concept-agnostic models and the variants of the concept-informed models.}
\label{tab:model_performance}
\centering
\begin{tabular}{lccc}
\toprule
 & \textbf{Agnostic} & \multicolumn{2}{c}{\textbf{Concept-informed}} \\
\cmidrule(lr){3-4}
\textbf{Dataset / metric} & -- & \textbf{Custom} & \textbf{DWT} \\
\midrule
\shortstack[l]{\textbf{OPPORTUNITY}\\ Weighted acc. (\%)} 
& 75.78 $\pm$ 3.57 & 76.87 $\pm$ 1.24 & 73.59 $\pm$ 1.45  \\
\shortstack[l]{\textbf{Turbofan}\\ RMSE (cycles)} 
& 20.87 $\pm$ 1.08 & 21.83 $\pm$ 1.38 & 22.45 $\pm$ 1.46  \\
\bottomrule
\end{tabular}
\end{table*}

\subsection{SHAP} 
The SHAP values for C-SHAP are calculated for the test samples. For the masking of concepts, the component masking, described in Section~\ref{sec:component_masking}, is applied using $N=100$ background samples from the training data. These samples were selected randomly, where for the classification task, the sampling was weighted per class. This results in 25 samples per class for the OPPORTUNITY dataset. The same background samples are used for all models trained on the same dataset, regardless of the model type.

The computational cost of SHAP scales exponentially with the number of features. This is why generally methods for estimation, such as KernelSHAP~\citep{lundberg_unified_2017}, are used to calculate the SHAP values. Since only a limited number of concepts are used, we calculate the SHAP values exactly.

\section{Results}
\label{sec:results}

In this section, we present and discuss the explanations generated by C-SHAP and relate these to human intuition. We show examples of local, sample-level, explanations and global explanations, where SHAP values are aggregated to estimate the influence of concepts across the datasets. Note that as a result of normalization of the data, the signals and decompositions shown in this section are dimensionless. 

\subsection{Local explanations}
\label{sec:res_anecdotal_local}
\begin{figure}[htp]
    \centering
    \begin{subfigure}{0.48\linewidth}
        \centering
        \includegraphics[width=\linewidth]{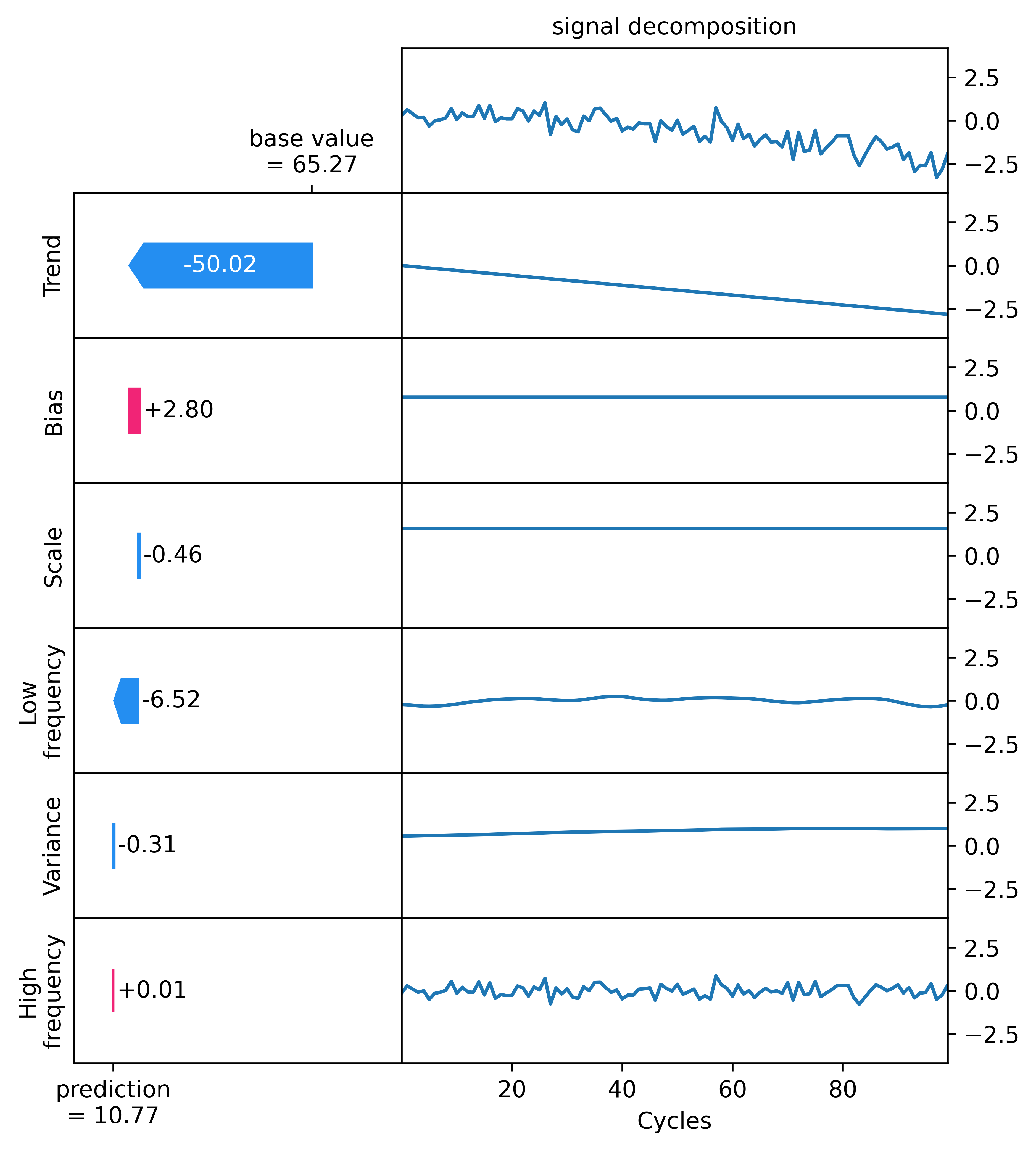}
        \caption{C-SHAP}
        \label{fig:local_turbo_cshap}
    \end{subfigure}%
    \begin{subfigure}{0.48\linewidth}
        \centering
        \includegraphics[width=\linewidth]{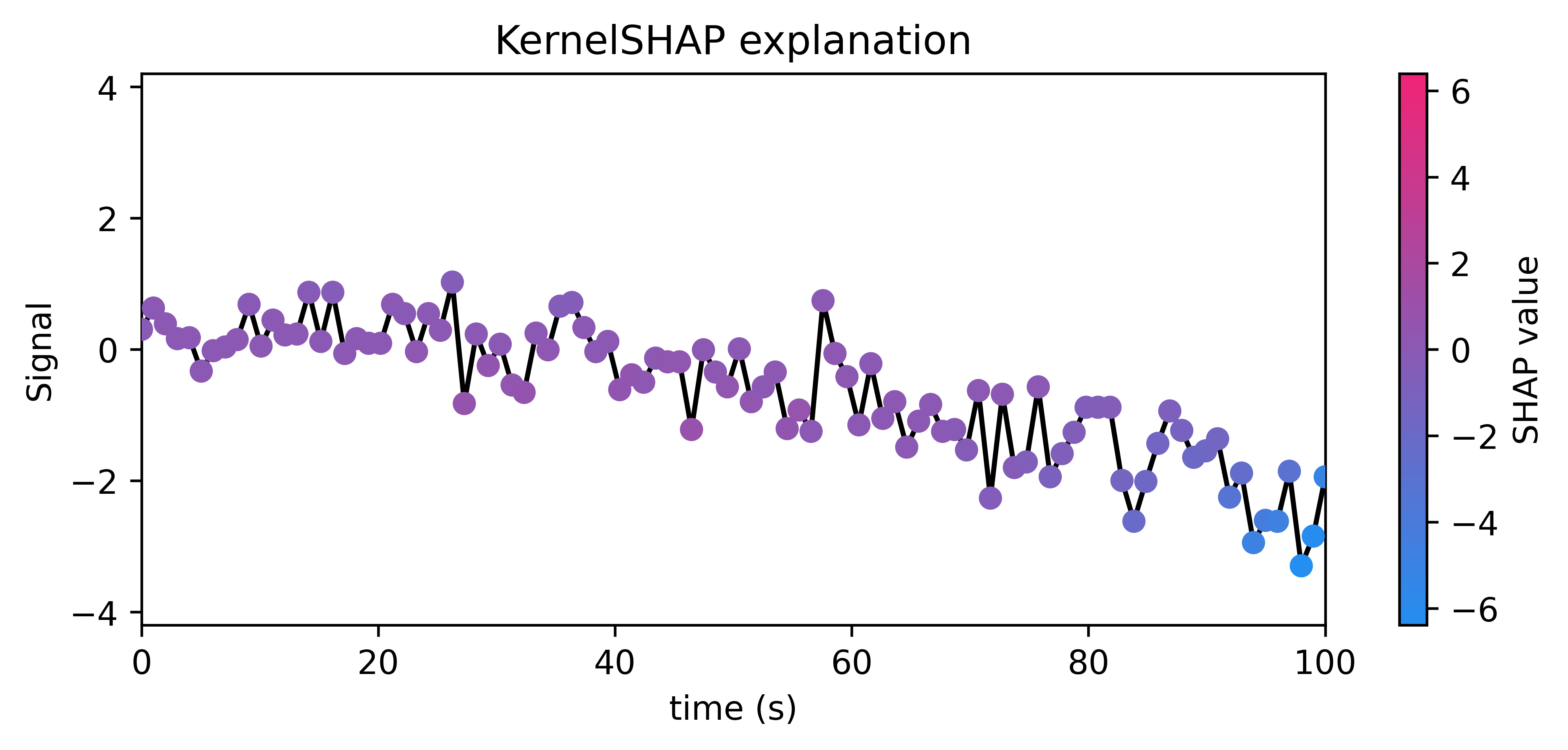}
        \caption{KernelSHAP}
        \label{fig:local_turbo_kernel}
    \end{subfigure}

    \caption{Local explanations using C-SHAP and KernelSHAP for a concept-agnostic model trained on the Turbofan dataset. The C-SHAP explanation has been generated using the custom decomposition.}
    \label{fig:turbo_local_cvskernel}
\end{figure}

In Figure~\ref{fig:turbo_local_cvskernel}, we show a C-SHAP explanation alongside an explanation generated using point-based KernelSHAP~\citep{lundberg_unified_2017}. The explanations are generated for a concept-agnostic model trained on the Turbofan data. It can be seen that, while C-SHAP clearly indicates the importance of `Trend' for the model's reasoning, KernelSHAP is not able to uncover this due to the global nature of the concept.

\begin{figure}[htp]
% \vspace{5mm}

    \centering
    \begin{subfigure}[t]{0.48\textwidth}
        \centering
        \includegraphics[width=\linewidth]{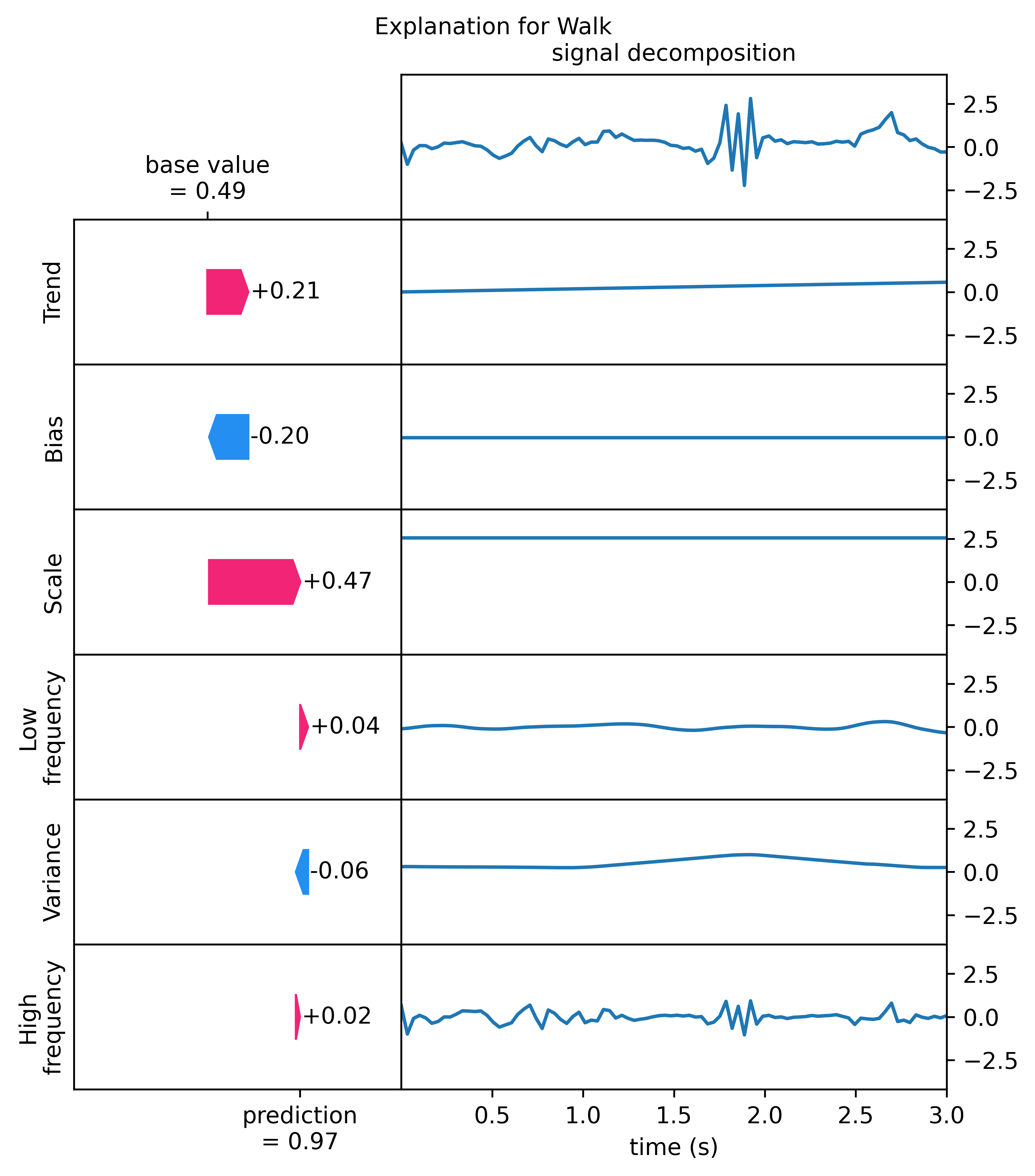}
        \caption{Custom decomposition}
    \end{subfigure}%
    \begin{subfigure}[t]{0.48\textwidth}
        \centering
        \includegraphics[width=\linewidth]{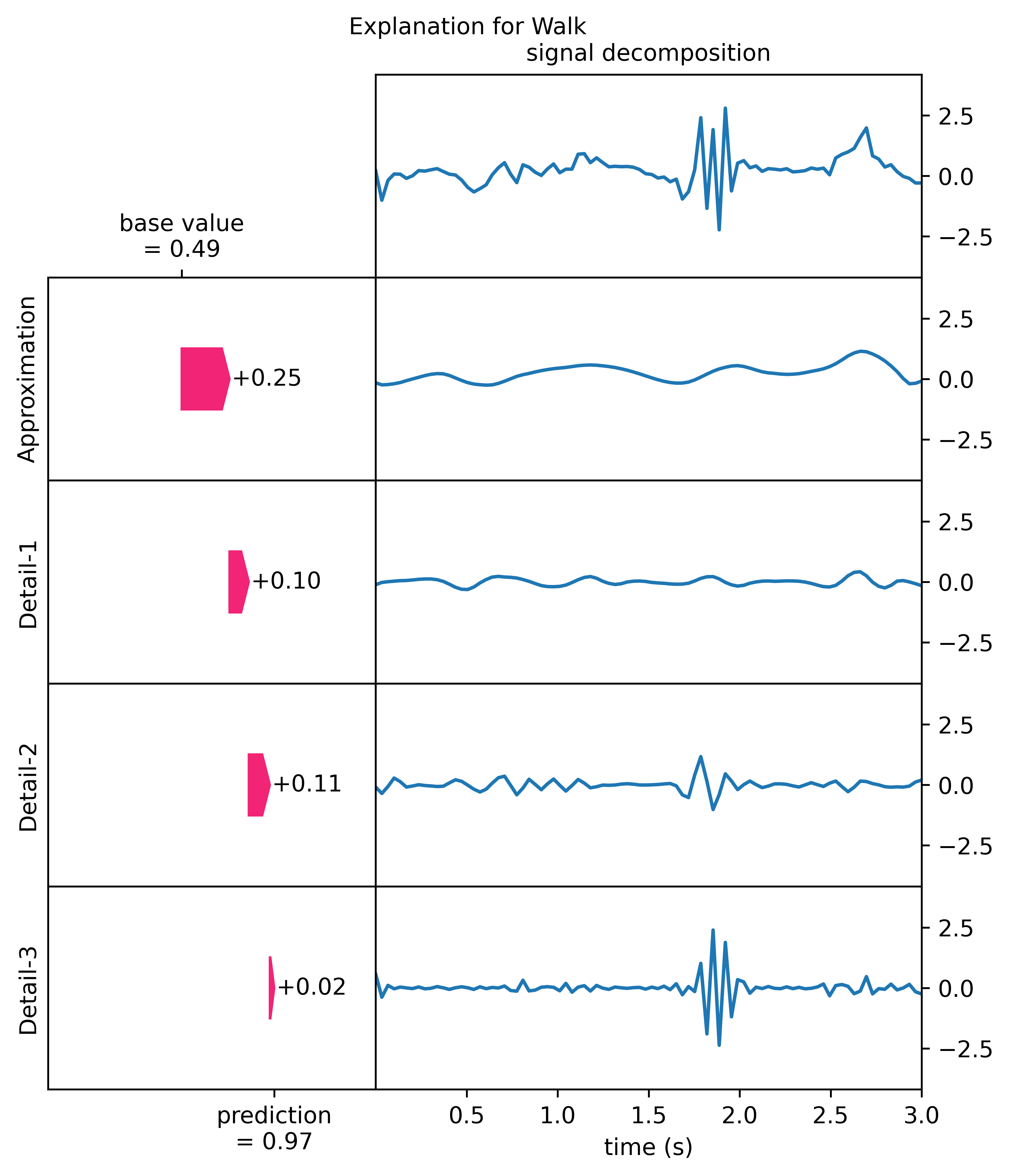}
        \caption{DWT}
    \end{subfigure}%

    \caption{Local explanations using the different decompositions, for a concept-agnostic model trained on the OPPORTUNITY dataset. Results are shown for the custom decomposition and Discrete Wavelet Transform.}
    \label{fig:opp_local}
\end{figure}

For a sample from the OPPORTUNITY dataset, local explanations generated using each of the decomposition methods are shown in Figure~\ref{fig:opp_local}. The explanations are generated for a concept-agnostic model.

The explanations in Figure~\ref{fig:opp_local} demonstrate how different decompositions can be used to generate explanations. From the explanation generated using the custom decomposition, we find `Scale' as the most influential concept, followed by `Trend' and `Bias'. The importance of `Scale' indicates that the high fluctuations are important for the model's reasoning. However, this is not due to the specific shape of the oscillation, since the `High frequency' concept is not considered important, nor is it the change of the fluctuations over time, since the `Variance' component is not considered important either. Rather, it is the extent of the amplitude that affects model reasoning.

For the explanations generated using the Discrete Wavelet Transform (DWT), the `Approximation' component is shown to be the most influential. Notably, the component capturing the high-frequency oscillation, `Detail-3', results in relatively low SHAP values. This is likely due to the additive nature of the DWT, which distributes the maximum amplitude over multiple components, rather than capturing it in an individual component. In our custom decomposition, `Scale' explicitly captures this behaviour.

\subsection{Global explanations}
In this section, we present global explanations generated using the custom decomposition for each of the datasets. We discuss which concepts were found to be most important and how these relate to human intuition.

\begin{figure}[htp]
    \centering

    \begin{subfigure}{0.48\linewidth}
        \centering
        \includegraphics[width=\linewidth]{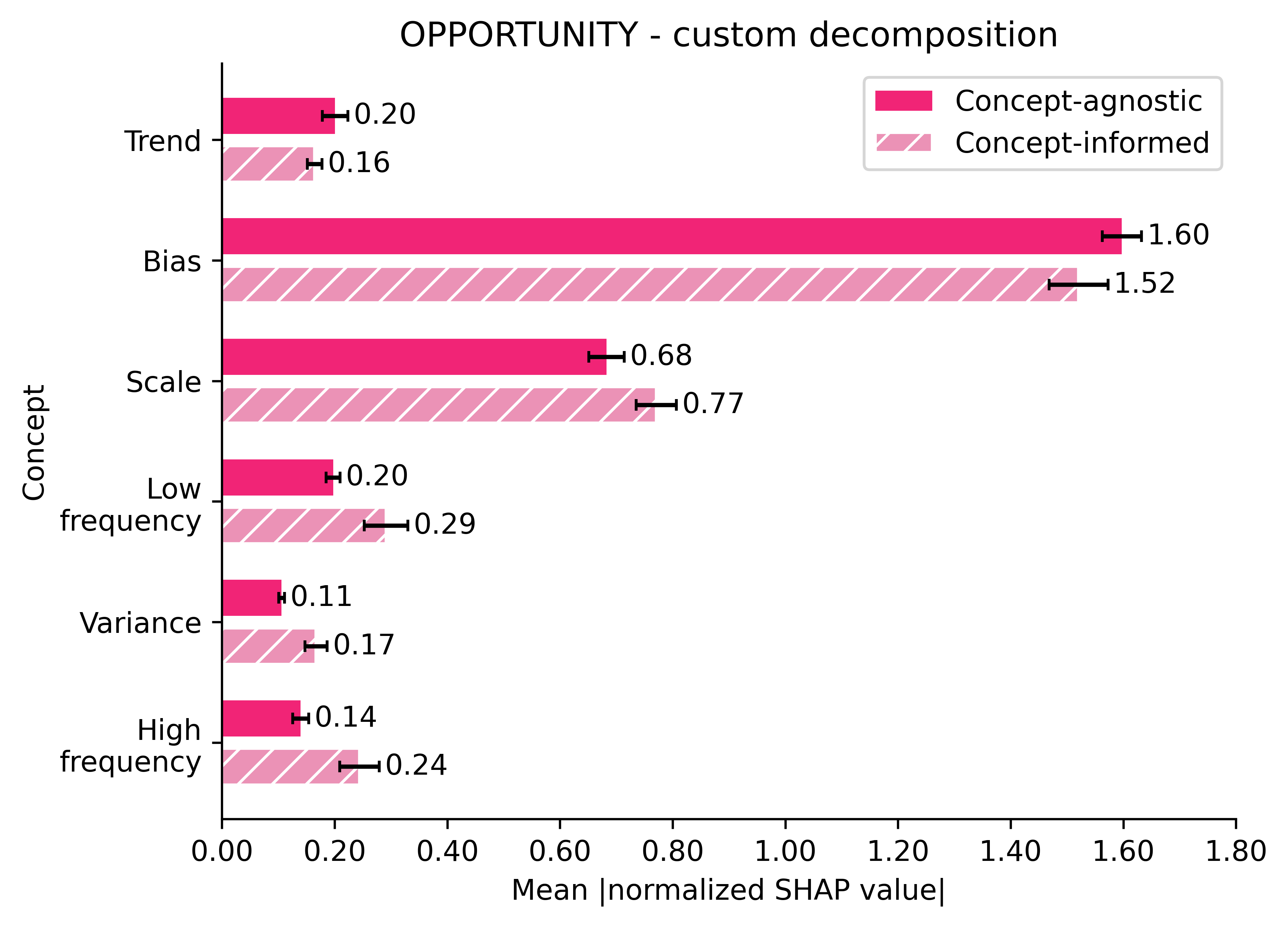}
        \caption{OPPORTUNITY}
    \end{subfigure}%
    \begin{subfigure}{0.48\linewidth}
        \centering
        \includegraphics[width=\linewidth]{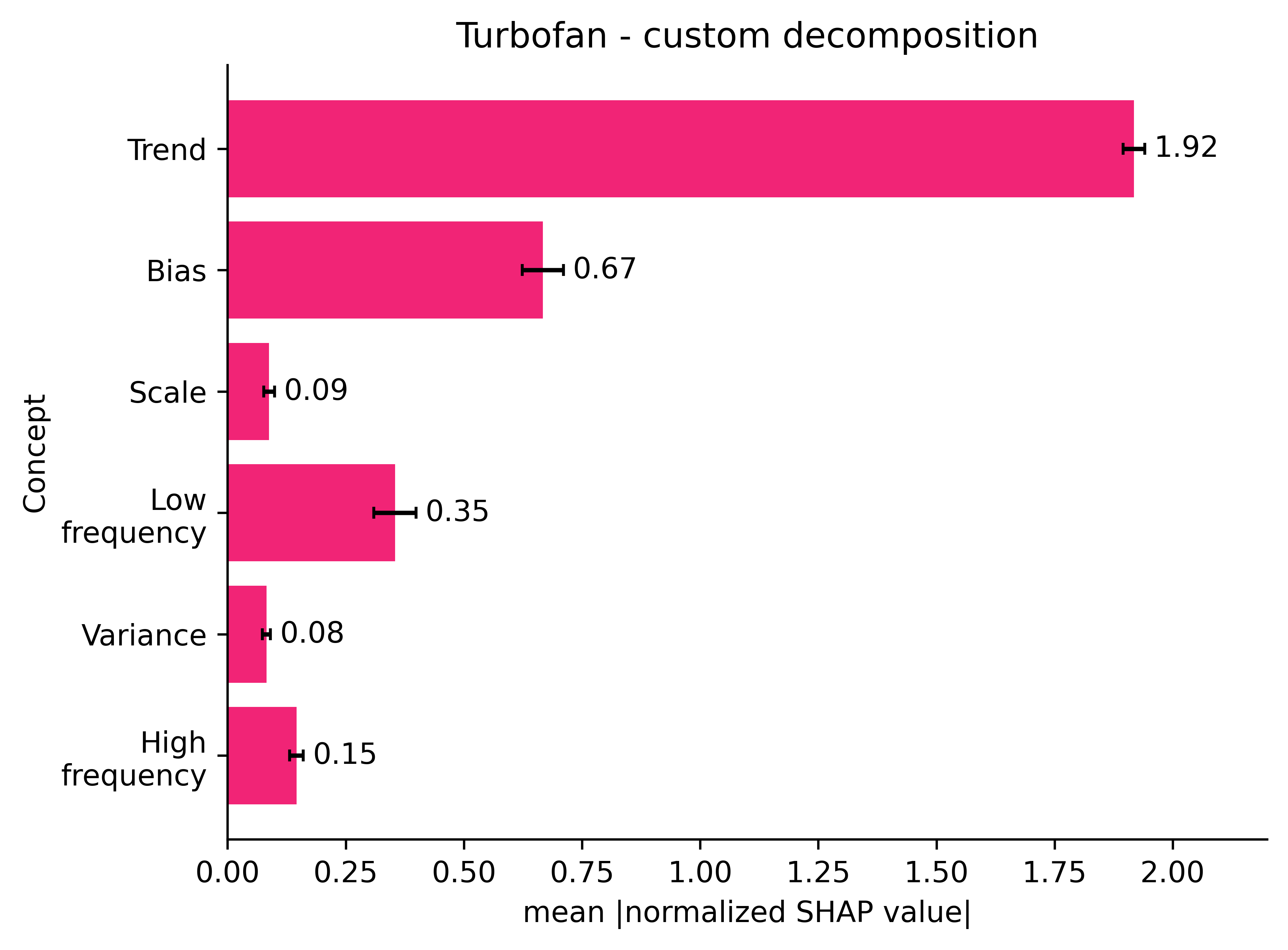}
        \caption{Turbofan}
    \end{subfigure}

    \caption{The absolute normalised SHAP values for models trained on the OPPORTUNITY and Turbofan datasets using the custom decomposition. Results are shown as the average and standard deviation over 10 runs. For the OPPORTUNITY dataset, results for the concept-agnostic and -informed models are included. For the Turbofan dataset, only results for the concept-agnostic models are shown.}
    \label{fig:global}

\end{figure}

For each of the models trained on the datasets, we calculated the mean absolute SHAP value over the test samples. The average and standard deviation of these values over the models are shown in Figure~\ref{fig:global}.

For the models trained on the OPPORTUNITY data, the models classify samples as `Stand', `Walk', `Sit' and `Lie' using accelerometer measurements from the subjects' right knee. The SHAP values indicate that the models mainly rely on the concepts `Bias' and `Scale'. We defined `Bias' as the average level of a signal, hence `Bias' represents the average angle of the subjects' knee. It seems reasonable that the model would use the angle of the knee to distinguish between the classes in the dataset. When a subject is standing, their knee is expected to be at a different angle, with respect to the normal axis of the accelerometer, than when they're sitting or lying. `Scale' captures the amplitude of the highest fluctuation in the signal. When a subject is walking, the maximum amount of movement in a window is expected to be higher than for the other classes. Hence, reliance on `Scale' seems reasonable as well. 

For the Turbofan models, the SHAP values identify `Trend' as the most influential concept. For these models, the pressure at the high-pressure compressor (HPC) outlet is used to predict the number of cycles the engine is still useful. The SHAP values indicate that the linear change in pressure is most indicative of remaining life.

\section{Discussion}

In our experiments, we applied C-SHAP to concept-informed and concept-agnostic models. While the performances and explanations we found seem comparable for the considered datasets and models, this does not need to hold for all applications. In literature, it has been shown that some decomposition-informed models reach higher performance levels~\citep{liu2013forecasting,yang2019hybrid}. Whether explicitly providing models with concept information increases performance might depend on the complexity of the task and the model architecture used.

For the concept-agnostic models, we limited our scope to models trained directly on the raw data. However, the `black box' concept-agnostic model in Figure~\ref{fig:cshap_approaches} could also include manual feature extraction. This allows the explicit use of features (concepts) during model training that may be beneficial for model performance, but which might not necessarily be interpretable by end users. Consider, for example, an application featuring a model trained using the DWT decomposition, but explained to lay users using the custom decomposition. This illustrates the flexibility and ease of application of the fully post-hoc approach.

The drawback of the flexible post-hoc approach for concept-agnostic models, is that it poses conditions on the concept construction. In this paper, we proposed that the construction algorithm needs to be invertible, see Condition~\ref{cond:invertible}. In this paper, we focus on concept construction through time series decomposition, which is invertible. In future research, it should be explored whether other concept construction methods exist that satisfy invertibility, and whether the proposed condition is strictly necessary. For the concept-informed models, we are not limited by this restriction. Any feature-extraction method can be considered as concept construction. For example, statistical features such as autoregression may be considered.

In this research, we defined concepts in terms of time series decomposition and considered two types of decomposition approaches. The decision for a decomposition is highly dependent on the application area for which an explanation is designed and its stakeholders. The custom decomposition might be easier to interpret for lay users. Consider, for example, the different explanations in Figure~\ref{fig:opp_local}. However, decompositions such as DWT may be more informative to experts in the field. Future research should explore the use of different decompositions in a range of domains. This would further validate the generalizability of C-SHAP. 

We presented the custom decomposition as an approach designed from a human-centred perspective. Its construction and implementation stemmed from application-driven considerations. However, its mathematical validity as a decomposition should be further proven. Additionally, to validate the interpretability of explanations generated using this decomposition, and other decompositions, C-SHAP should be evaluated through user study evaluations with domain experts and lay users. 

The explanations generated using C-SHAP rely on the selected concept construction method and its hyperparameters. Changing this order may have significant effects on the generated explanations. Furthermore, a filtering threshold and a wavelet are selected to extract the `Low frequency' component, as well as a window size to extract the `Variance' component. For this paper, we manually selected the hyperparameters through visual inspection. The effect of changes in hyperparameters on the explanations should be evaluated in future research. Additionally, methods to automatically select or optimise the hyperparameters could be explored.

\section{Conclusion}
\label{sec:conclusion}

Concept-based explanations can uncover high-level patterns used in the internal reasoning of black-box models. Furthermore, they may provide explanations in terms matching human understanding. In this article, we presented C-SHAP as an approach for concept-based explanations for time series. We defined two approaches for C-SHAP, one fully post-hoc approach, applicable to any time series model, and one partially ante-hoc approach, in which the model is informed by the concepts under inspection. C-SHAP offers custom control over concept selection. To demonstrate how C-SHAP can be applied, we presented an implementation using time series decomposition. 

To bridge the gap between theory and application, we applied C-SHAP to a Human Activity Recognition and a predictive maintenance use case. We used C-SHAP to uncover the importance of the bias and scale of signals in models trained for a locomotion classification task and the importance of trend in a predictive maintenance regression task.

We emphasise the generalizability of our methodology to other time series use cases and concept construction approaches. In future work, C-SHAP should be applied to a wide range of domains to explore the effectiveness of different concept construction approaches. Furthermore, to further validate its interpretability, C-SHAP should be evaluated among end users.

\section*{Acknowledgements} 
This publication is part of the project ZORRO with project number KICH1.ST02.21.003 of the research programme Key Enabling Technologies (KIC) which is partly financed by the Dutch Research Council (NWO). This research is part of the SPRONG DEMAND. This research is partly financed by Taskforce for Applied Research SIA, part of the Dutch Research Council (NWO).

\bibliographystyle{plainnat}
\bibliography{bibliograpy}

\end{document}